\documentclass[10pt,twocolumn,letterpaper]{article}

\usepackage{iccv}
\usepackage{times}
\usepackage{epsfig}
\usepackage{graphicx}
\usepackage{amsmath}
\usepackage{amssymb}

\usepackage{graphicx}
\usepackage{booktabs}
\usepackage{multirow}
\usepackage{tabu}
\usepackage{subcaption}

\usepackage{csquotes}
\usepackage{makecell}
\usepackage{authblk}

\usepackage[dvipsnames]{xcolor}
\usepackage{color, colortbl}
\definecolor{citecolor}{HTML}{0071bc}
\definecolor{tabhighlight}{HTML}{e5e5e5}

\newcommand{\std}[1]{\tiny{$\pm$#1}}

\usepackage[breaklinks=true,bookmarks=false]{hyperref}

\iccvfinalcopy 

\def\iccvPaperID{5737} 

\ificcvfinal\fi

\begin{document}

\title{Why Is Prompt Tuning for Vision-Language Models Robust to Noisy Labels?}

\author{Cheng-En Wu$^1$\thanks{Work done during an internship at ByteDance Inc.} \;\, Yu Tian$^2$ \;\, Haichao Yu$^2$ \;\, Heng Wang$^2$\;\, Pedro Morgado$^1$ \\ Yu Hen Hu$^1$\;\, Linjie Yang$^2$ \\
\vspace{0.5em}
$^1$University of Wisconsin-Madison \,\; $^2$ByteDance Inc. \\
{\tt\small \{cwu356, pmorgado, yhhu\}@wisc.edu \,\{yutian.yt, haichaoyu, heng.wang, linjie.yang\}@bytedance.com

}\vspace{-1em}
}

\maketitle
\ificcvfinal\thispagestyle{empty}\fi

\begin{abstract}
Vision-language models such as CLIP~\cite{Radford2021CLIP} learn a generic text-image embedding from large-scale training data. A vision-language model can be adapted to a new classification task through few-shot prompt tuning. We find that such a prompt tuning process is highly robust to label noises. This intrigues us to study the key reasons contributing to the robustness of the prompt tuning paradigm. We conducted extensive experiments to explore this property and find the key factors are: 
1) the fixed classname tokens provide a strong regularization to the optimization of the model, reducing gradients induced by the noisy samples; 
2) the powerful pre-trained image-text embedding that is learned from diverse and generic web data provides strong prior knowledge for image classification. 
Further, we demonstrate that noisy zero-shot predictions from CLIP can be used to tune its own prompt, significantly enhancing prediction accuracy in the unsupervised setting. The code is available at \url{https://github.com/CEWu/PTNL}.
\end{abstract}

\section{Introduction}
\label{sec:intro}
\begin{figure}[ht]
\centering
 \begin{subfigure}{0.8\columnwidth}
     \includegraphics[width=\columnwidth]{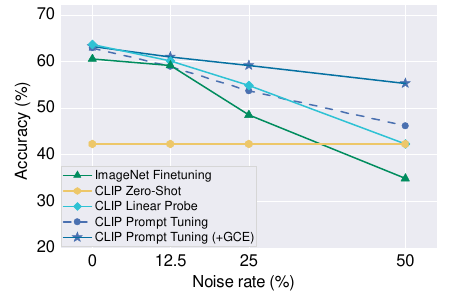}
     \caption{DTD}
     \label{fig:CLIP-PTvsRN5-FT_DTD}
 \end{subfigure}
 \hfill
 \begin{subfigure}{0.8\columnwidth}
     \includegraphics[width=\columnwidth]{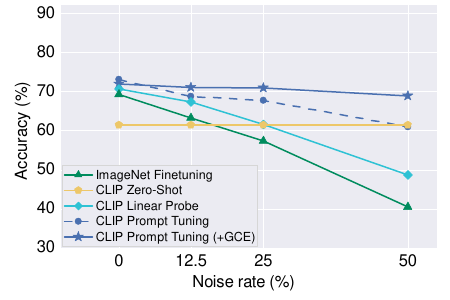}
     \caption{UCF101}
     \label{fig:CLIP-PTvsRN5-FT_ucf101}
 \end{subfigure}
 \caption{Comparison with transfer learning approaches on two datasets with training labels that have incremental noisy rates. ImageNet Finetuning is finetuning pre-trained model on ImageNet. For the CLIP pre-trained model, Prompt Tuning is much more robust to the Linear Probe manner. By combining the generalized cross-entropy (GCE)~\cite{GenXEnt}, we further improve the robustness of Prompt Tuning to noisy labels. ResNet-50 is used for all approaches as their image encoders.}
 \label{fig:CLIP-PTvsRN5-FT}
\end{figure}

Large-scale vision-language models such as CLIP~\cite{Radford2021CLIP}, ALIGN~\cite{jia2021_align}, and CoCa~\cite{yu2022coca} are transforming how we learn and interact with visual representations. Since these models learn to align the representations of a broad set of natural images with their textual descriptions, they have shown an exceptional ability to solve a wide range of tasks in a data-efficient manner. For example, using the pre-trained text encoder, one can obtain a set of class embeddings by encoding a canonical sentence such as ``A photo of a $<$CLS$>$'' and use them to recognize objects without a labeled dataset. While promising, Zhou et al.~\cite{Zhou2022CoOp} showed that these human-defined sentences (also known as class prompts) can be unstable, with seemingly equivalent descriptions leading to different predictions. To address this issue, researchers have focused on prompt tuning~\cite{Zhou2022CoOp}, where a learnable prompt is learned from a small target dataset by back-propagation. Since only the prompt needs to be trained, this framework is very data-efficient. As a result, prompt-tuning has gained popularity for adapting vision-language models to downstream tasks like few-shot learning~\cite{Zhou2022CoOp,zhou2022CoCoOp}, continual learning~\cite{wang2022continual}, and object segmentation~\cite{rao2022denseclip}.

While prompt tuning has proven effective when training on downstream tasks with accurately annotated datasets, their robustness to noisy labels has been neglected. Since the quality of annotations for many applications can be low, learning with noisy labels is critical to solving real-world problems. In this work, we demonstrate that prompt tuning is robust to noisy labels, and investigate the mechanisms that enable this robustness. We hypothesize that the joint text and image embeddings of vision-language models can provide a well defined structure to the classification space (e.g., which classes are most similar and most distinct from each other). This model-informed structure compensates for
the degradation of the structure present in the data due to label noise. To verify this hypothesis, we conducted extensive experiments to study the impact of each component of a prompt tuning task with noisy labeled data. 
Beyond the robustness conferred by the structured label space, we show that this robustness can be further enhanced when the learnable prompts are trained using a robust loss function that mitigates the impact of outliers. Our study has revealed several interesting findings.

First, the classification performance obtained by tuning the prompt through a pre-trained CLIP model is significantly more robust to noisy labels than the traditional fine-tuning or linear probing paradigms (see Figure~\ref{fig:CLIP-PTvsRN5-FT}).
The robustness of prompt tuning is evident not only due to their smaller performance degradation with higher noise rates, but also due to its ability to diminish the gradients induced by noisy samples.
Second, while priming each class with a shared learnable prompt is necessary for adaptation, ensuring that the class name remains in the prompt strongly regularizes the class embeddings and prevents overfitting to the noisy labels. 
Finally, we demonstrate the benefits of this robustness by showing that CLIP zero-shot (noisy) predictions can be used to tune its own prompt, and significantly enhance CLIP prediction accuracy. In fact, we show that, instead of focusing on samples with confident predictions (as proposed in prior unsupervised prompt tuning approaches~\cite{Huang2022UPL}), prompt tuning benefits more from an increased diversity of training samples as it can tolerate the noisier predictions associated with them.

The main contributions of our work are as follows:
\begin{itemize}
    \setlength\itemsep{0.1em}
    \item We demonstrate that prompt tuning for pre-trained vision-language models (e.g., CLIP) is more robust to noisy labels than traditional transfer learning approaches, such as model fine-tuning and linear probes.
    \item We further demonstrate that prompt tuning robustness can be further enhanced through the use of a robust training objective.
    \item We conduct an extensive analysis on why prompt tuning is robust to noisy labels to discover which components contribute the most to its robustness.
    \item Motivated by this property, we propose a simple yet effective method for unsupervised prompt tuning, showing that randomly selected noisy pseudo labels can be effectively used to enhance CLIP zero-shot performance. The proposed robust prompt tuning outperformed prior work~\cite{Huang2022UPL} on a variety of datasets, even though noisier pseudo-labels are used for self-training.
\end{itemize}

\section{Related Work}
\label{sec:r_work}

\noindent \textbf{Prompt tuning for Vision-Language models.} Over the past few years, there has been huge progress in Vision-Language Pre-Trained Models (VL-PTMs)~\cite{Radford2021CLIP,jia2021_align,yao2021filip,yu2022coca}. CLIP~\cite{Radford2021CLIP} is considered a representative model of VL-PTMs. Unlike the conventional, finetuning paradigm, CLIP applies prompt engineering to incorporate the category information in the text input such that its pre-trained model can adapt to various image classification tasks without further training. However, the design of a proper prompt is challenging and requires heuristics. CoOp~\cite{Zhou2022CoOp} introduces learnable prompts optimized on target datasets to address this problem. 
To further extend the generalization of CoOp, CoCoOp~\cite{zhou2022CoCoOp} introduces a lightweight network to add additional information from image inputs into learnable prompts. CoOp has also faced criticism for disregarding the diversity of visual representations. In contrast, ProDA~\cite{lu2022prompt_dist} tackles this issue by utilizing diverse prompts to capture the distribution of varying visual representations.

In contrast to the supervised tuning methods above, UPL~\cite{Huang2022UPL} proposes a framework to perform prompt tuning without labeled data. TPT~\cite{shu2022tpt} achieves zero-shot transfer by dynamically adjusting prompts using only a single test sample.

In addition to downstream tasks for image classification,  recent works have applied prompt tuning on CLIP to various computer vision tasks such as object detection~\cite{rao2022denseclip,du2022Detectiong}, video understanding~\cite{li2022bridge,ju2022prompting_video}, and multi-label recognition~\cite{sun2022dualcoop}. These works reveal the further potential of prompt tuning. 

\noindent \textbf{Label noise-robust learning.} Deep neural networks (DNNs) have been well-studied for classification tasks without label noises. However, if the training data contains label noise, DNNs would easily overfit to the noisy labels~\cite{Zhang2017Understanding}. To overcome this issue, several works have attempted to improve the noise robustness of DNNs by approaches including robust losses that tolerate noisy labels~\cite{ghosh2017robust_loss, GenXEnt,wang2019symmetric,ma2020normalized}, loss correction approaches that estimate a transition matrix to correct the predictions~\cite{patrini2017making,hendrycks2018using,chang2017active,bootstrapping,arazo2019unsupervised,ma2018dimensionality,song2019selfie,yi2019probabilistic,yao2020dual}, meta-learning frameworks that learn to correct the label noise in training examples~\cite{li2017learning,ren2018learning,li2019learning,shu2019meta,zhang2020distilling,zheng2021meta} and regularization techniques that are customized to lower the negative impact of noise~\cite{mixup,smoothing,hendrycks2019using, xia2020robust}. 

In this work, we demonstrate that prompt tuning on CLIP naturally holds powerful noise robustness. We explore the key factors behind such robustness and show its application on unsupervised prompt tuning.

\begin{figure*}[t]
 \begin{subfigure}{0.5\columnwidth}
     \includegraphics[width=\columnwidth]{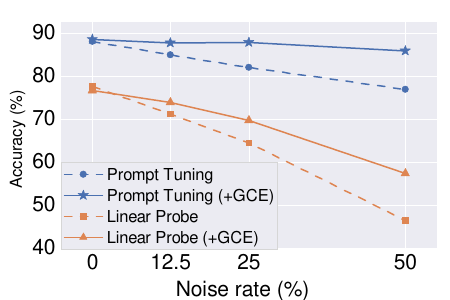}
     \caption{OxfordPets}
     \label{fig:GCE_oxfordpets}
 \end{subfigure}
 \hfill
 \begin{subfigure}{0.5\columnwidth}
     \includegraphics[width=\columnwidth]{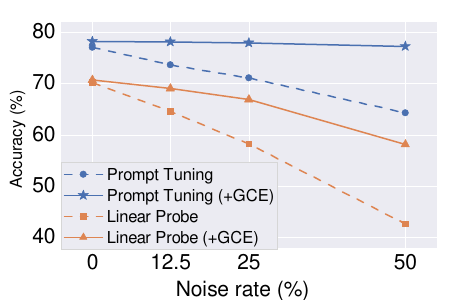}
     \caption{Food101}
     \label{fig:GCE_food101}
 \end{subfigure}
  \hfill
  \begin{subfigure}{0.5\columnwidth}
     \includegraphics[width=\columnwidth]{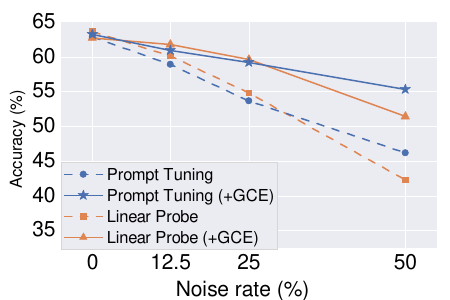}
     \caption{DTD}
     \label{fig:GCE_DTD}
 \end{subfigure}
  \hfill
  \begin{subfigure}{0.5\columnwidth}
     \includegraphics[width=\columnwidth]{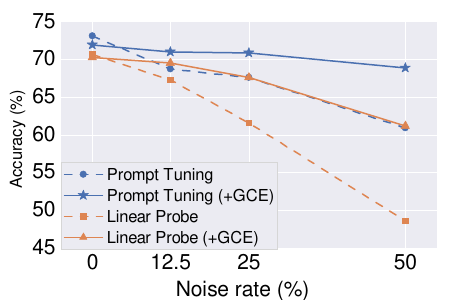}
     \caption{UCF101}
     \label{fig:GCE_ucf101}
 \end{subfigure}
 \caption{Incorporating the generalized cross-entropy (GCE)~\cite{GenXEnt} loss with Prompt Tuning and Linear Probe methods, originally trained using cross-entropy, can enhance their noise robustness. At high noise rates, Prompt Tuning with GCE outperforms other methods by a significant margin across four datasets.}
 \label{fig:GCE_PTvsLP}
\end{figure*}

\section{Prompt Tuning}
\label{sec:method}
CLIP~\cite{Radford2021CLIP} can perform zero-shot transfer by prompt engineering -- the practice of designing text inputs for downstream tasks. Specifically, in the case of image classification, a normalized image embedding $\mathbf{f}^v$ is obtained by passing an image $\mathbf{x}$ through CLIP's visual encoder, and a set of normalized class embeddings $\{\mathbf{f}^t_{i}\}^K_{i=1}$ by feeding template prompts of the form ``A photo of a $<$CLS$>$'' into CLIP's text encoder.
The class posterior is then estimated as
\begin{equation}
Pr(y=i|\mathbf{x}) =  \frac{\exp \Bigl(\mathtt{sim}(\mathbf{f}^v, \mathbf{f}^t_i) / \tau \Bigr) }{\sum_{j=1}^{K} \exp \Bigl(\mathtt{sim}(\mathbf{f}^v, \mathbf{f}^t_j)/\tau\Bigr)},
\label{eq:cos_sim}
\end{equation}
where $\tau$ is a temperature factor learned by CLIP and $\mathtt{sim}$ denotes cosine similarity.

\paragraph{Prompt Tuning}
Although CLIP is capable of zero-shot transfer, its performance is sensitive to designed text prompts. 
To avoid the need for hand-crafted prompts and improve transfer performance, CoOp~\cite{Zhou2022CoOp} showed that text prompts can be replaced with continuous soft prompts that can be optimized on a target dataset. 
Specifically, the name of a class $c$ is first converted into a classname embedding $\mathbf{w}_c \in \mathbb{R}^{d}$ and prepended with a sequence of $M$ learnable tokens $\mathbf{p}_m \in \mathbb{R}^{d}$ shared across all classes. The full prompt $\mathrm{P_c}=[\mathbf{p}_1,\mathbf{p}_2,...\mathbf{p}_M,\mathbf{w}_c]$ for each class $c$ is then processed by CLIP's text encoder to compute the corresponding text embedding $\mathbf{f}^t_c$, and the class posteriors $Pr(y=i|\mathbf{x})$ are obtained once again through Eq.~\ref{eq:cos_sim}.
To adapt the prompt to the target dataset, CoOp~\cite{Zhou2022CoOp} optimizes the shared learnable tokens $\mathbf{p}_1,\mathbf{p}_2,...\mathbf{p}_M$ on a small labeled dataset $\mathcal{D}=\{({\bf x}_i, c_i)_{i=1}^N\}$ to minimize the cross-entropy loss
\begin{equation}
    \mathcal{L}_{CE} = -\mathbb{E}_{(\mathbf{x}, c)\in\mathcal{D}} \left[\log 
    Pr(y=c|\mathbf{x})\right].
    \label{eq:lossCE}
\end{equation}

\paragraph{Robust Prompt Tuning}
In this work, we show that the prompt tuning framework~\cite{Zhou2022CoOp}, describe above, displays surprising robustness to noisy labels. However, this robustness can be further enhanced by optimizing the learnable prompts using the generalized cross-entropy (GCE) loss~\cite{GenXEnt}, a robust generalization of cross-entropy loss. Formally, the GCE loss is defined as
\begin{equation}
    \mathcal{L}_{GCE} = \mathbb{E}_{(\mathbf{x}, c)\in\mathcal{D}} \left[
        \frac{1 - Pr(y=c|\mathbf{x})^q}{q}
    \right].
    \label{eq:lossGCE}
\end{equation}
As shown in~\cite{GenXEnt}, GCE is equivalent to the standard cross-entropy loss of Eq.~\ref{eq:lossCE} when $x \to 0$, and equivalent to the (robust) mean absolute error (MAE) loss ${\|1 - Pr(y=c|\mathbf{x})\|_1}$ when $q = 1$. The hyper-parameter $q$ can therefore control the tradeoff between the highly robust but less performing MAE loss and the less robust but highly performing CE loss. While the optimal value for $q$ could be adjusted to the amount of noise by cross-validation, we found that $q=0.7$ lead to overall good performance across several experimental settings.

\begin{figure*}[t]
  \begin{center}
    \includegraphics[width=0.85\textwidth]{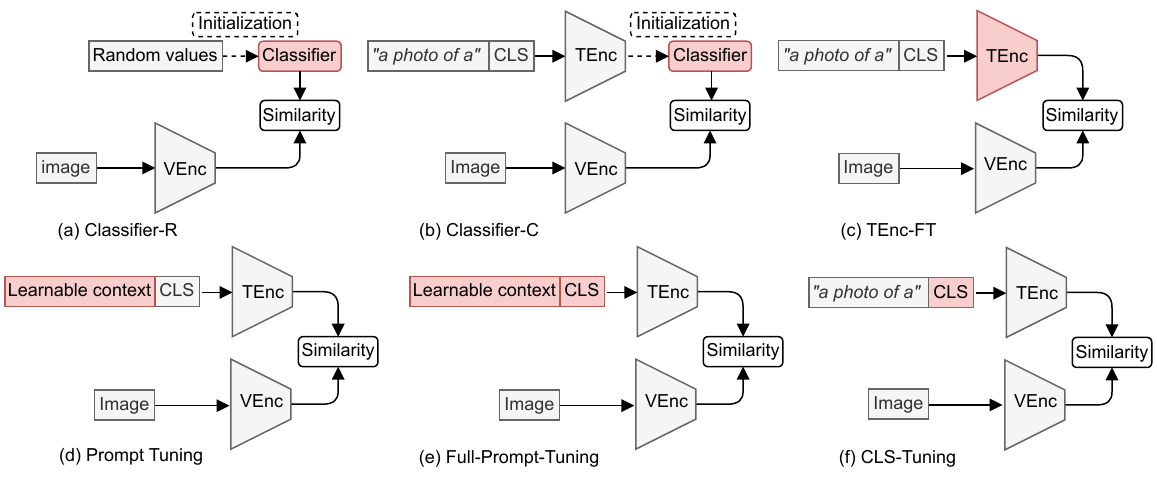}
  \end{center}
  \caption{Illustration of different structures for studying the effect of image and text encoders on prompt tuning and prompt design. The blocks highlighted in red are to be trained, while those highlighted in gray are to be frozen.}
  \label{fig:PT_ablation_TEnc_Prompt}
\end{figure*}

\section{Analysis of Prompt Tuning with Label Noise}
\label{sec:exp-robustness}
Methods based on prompt tuning for CLIP~\cite{Radford2021CLIP} have been shown to be effective in few-shot learning~\cite{Zhou2022CoOp,zhou2022CoCoOp}. However, these methods have been studied on datasets with perfect labels. It remains unknown how prompt tuning performs under label noise. We explore this practical training setting and present our key findings.

\subsection{Experimental Settings}

\noindent\textbf{Datasets.} We conduct in-depth studies on a diverse set of visual tasks, including generic object classification, fine-grained recognition, action recognition, and texture identification. We conduct our experimental analysis on eight datasets, OxfordPets~\cite{parkhi2012pets}, Food101~\cite{bossard2014food}, DTD~\cite{cimpoi2014DTD}, UCF101~\cite{soomro2012ucf101}, Flowers102~\cite{nilsback2008flower}, FGVCAircraf~\cite{maji2013aircraft}, Caltech101~\cite{fei2004cal101} and ImageNet~\cite{recht2019imagenet}.
Since one of the main benefits of prompt tuning is its data efficiency~\cite{Huang2022UPL}, we focus our studies on a 16-shot image classification problem, i.e.~for each dataset, we select 16 images per class as our training set. To examine the effect of noise in prompt tuning, we randomly perturb training labels with different levels of noise rate (12.5\%, 25\%, and 50\%). Unless otherwise specified, noisy labels are drawn uniformly at random from other categories of the dataset. We report average results over four runs with different training sets in all experiments.

\noindent\textbf{Backbone.} We adopt pre-trained CLIP models, namely using the 63M parameter text Transformer~\cite{vaswani2017attention} as the text encoder, and either a ResNet-50~\cite{he2016resnet} or a ViT-B/32~\cite{dosovitskiy2020ViT} as the visual encoder. Following CoOp~\cite{zhou2022CoCoOp}, we use 16 learnable tokens in each prompt shared across all categories. 

\noindent\textbf{Optimization.} Models are trained with a batch size of 32 for 50 epochs, using stochastic gradient descent (SGD) with momentum of 0.9 and an initial learning rate of 0.002, annealed to zero with a cosine decay schedule.

\subsection{Prompt Tuning Is Robust to Noisy Labels}
\label{sec:method_noise}

The core observation of this paper is that prompt tuning vision-language models, such as CLIP, is surprisingly robust to noisy labels.
This can be observed by comparing prompt tuning for CLIP with two traditional transfer learning approaches: 1) training a linear classifier on CLIP's visual representations (denoted CLIP Linear Probe); and 2) fine-tuning the same visual backbone pre-trained on ImageNet. 
The results on two datasets, DTD and UCF101, are shown in Figure~\ref{fig:CLIP-PTvsRN5-FT} (a) and (b), respectively. As can be seen, although linear probes and fine-tuning achieve competitive performance with perfectly labeled data (0\% noise rate), both procedures suffer from a significant accuracy drop with higher noise rates of 25\% and 50\%.
This result shows that prompt tuning is naturally more resistant to noisy labels than the alternatives. We show nevertheless that its robustness can be further enhanced by training the prompt using the robust generalized cross-entropy loss (denoted CLIP Prompt Tuning (GCE) in Figure~\ref{fig:CLIP-PTvsRN5-FT}). As can be seen, when combining Prompt Tuning and GCE, the model's performance remains highly competitive, even for noise rates as high as 50\%.
Furthermore, we observe that this robustness stems from the combination between Prompt Tuning and GCE, and not from GCE alone. This can be seen in Figure~\ref{fig:GCE_PTvsLP}, which depicts the noise robustness of Prompt Tuning and Linear Probes both trained under the cross-entropy and GCE losses on four datasets. While the robustness of the linear probe also improves with a GCE loss, the performance drop at high noise rates is significantly smaller when learning through prompt tuning.

Now that we have established the noise robustness of prompt tuning, the remainder of this Section is dedicated to providing intuitions and experimental analysis to answer the why question.
\begin{displayquote}
\textbf{\textit{Question:}} \textit{Why is prompt tuning for CLIP-like vision-language models more robust than traditional transfer learning against noisy labels?}
\end{displayquote}

\setlength{\tabcolsep}{3pt}
\begin{table}[t]
\small
\centering
\begin{tabular}{cl|cccccc}
\toprule
\multirow{2}{*}{\bf Dataset} &
\multirow{2}{*}{\begin{tabular}[c]{@{}c@{}} \bf Method \end{tabular}}
&\multicolumn{4}{c}{\bf Noise rate} \\ 
\multicolumn{2}{l|}{}                                 
& 0 & 12.5 & 25 & 50 \\ \cline{3-6} \hline

& Classifier-R
& 74.82
& 64.10
& 55.96
& 36.63 \\
& Classifier-C
& 81.47
& 70.29
& 61.87
& 44.21\\
& TEnc-FT
& 84.38
& 70.73
& 61.11
& 41.21 \\
\rowcolor{tabhighlight}
\multirow{-4}{*}{\cellcolor{white} OxfordPets} 
& Prompt Tuning
& \textbf{87.89}
& \textbf{84.62}
& \textbf{81.20}
& \textbf{73.13} \\ \hline

& Classifier-R
& 63.80
& 54.66
& 46.23
& 28.97 \\
& Classifier-C
& 69.36
& 60.46
& 51.85
& 34.37\\
& TEnc-FT
& 71.30
& 61.60
& 52.64
& 34.74 \\
\rowcolor{tabhighlight}
\multirow{-4}{*}{\cellcolor{white} Food101} 
& Prompt Tuning
& \textbf{76.99}
& \textbf{73.63}
& \textbf{71.07}
& \textbf{64.30} \\ \hline

& Classifier-R
& 48.02
& 44.30
& 40.32
& 30.10 \\
& Classifier-C
& \textbf{63.83}
& 57.14
& 50.36
& 34.86\\
& TEnc-FT
& 63.61
& 55.47
& 48.21
& 33.12 \\
\rowcolor{tabhighlight}
\multirow{-4}{*}{\cellcolor{white} DTD} 
& Prompt Tuning
& 62.86
& \textbf{58.90}
& \textbf{53.62}
& \textbf{46.19} \\ \hline

& Classifier-R
& 67.16
& 58.33
& 50.34
& 31.07 \\
& Classifier-C
& 71.87
& 64.12
& 54.79
& 38.01\\
& TEnc-FT
& \textbf{73.74}
& 64.52
& 56.10
& 37.88 \\
\rowcolor{tabhighlight}
\multirow{-4}{*}{\cellcolor{white} UCF101} 
& Prompt Tuning
& 73.12
& \textbf{68.73}
& \textbf{67.66}
& \textbf{60.93} \\ 

\bottomrule
\end{tabular}
\caption{Comparison of transfer performance at incremental noise rates between different variants.}
\label{tbl:classifier_comp}
\end{table}

\subsection{Robustness Attribution}
\label{sec:reason}
To answer this question, we begin by analyzing two key components of CLIP in isolation, namely the generated class embeddings and learnable prompts. 

\noindent \textbf{Pre-trained CLIP generates effective class embeddings.} 
We first analyse the impact of the class embeddings generated by the CLIP text encoder. To this end, in addition to the class embeddings generated through prompt tuning, we assess the noise robustness of three different models:
\begin{description}
    \setlength\itemsep{0.05em}
    \item[Classifier-R] Trains a linear probe on the output of CLIP's pre-trained visual encoder. The class embeddings (i.e., the classifier weights) are initialized at \emph{random}, and learned without constrains. See Figure~\ref{fig:PT_ablation_TEnc_Prompt} (a).
    \item[Classifier-C] Similar to Classifier-R, but the classifier weights are initialized using the text embeddings $\mathbf{f}_c^t$ obtained from CLIP's pre-trained text encoder for the handcrafted prompt. Note that Classifier-C only uses the CLIP text encoder for initializing its weights. See Figure~\ref{fig:PT_ablation_TEnc_Prompt} (b).
    \item[TEnc-FT] Trains a CLIP classifier, by associating the image embedding ${\bf f}^v$ with the CLIP text embedding ${\bf f}^t$ of the correct class through the posterior of eq.~(\ref{eq:cos_sim}).
    In this case, the entire CLIP text encoder is \emph{fine-tuned} on an hand-crafted prompt of the form ``A photo of a $<$CLS$>$''.
    See Figure~\ref{fig:PT_ablation_TEnc_Prompt} (c). 
\end{description}

\setlength{\tabcolsep}{2pt}
\begin{table}[t]
\small
\centering
\begin{tabular}{cl|cccccc}
\toprule
\multirow{2}{*}{\bf Dataset} &
\multirow{2}{*}{\begin{tabular}[c]{@{}c@{}} \bf Method \end{tabular}}
&\multicolumn{4}{c}{\bf Noise rate} \\ 
\multicolumn{2}{l|}{}                                 
& 0 & 12.5 
& 25
& 50 \\ \cline{3-6} \hline

& Full-Prompt-Tuning
& 85.39
& 74.00
& 68.66
& 50.50\\
& CLS-Tuning
& 85.04
& 77.02
& 71.03
& 53.15 \\
\rowcolor{tabhighlight}
\multirow{-3}{*}{\cellcolor{white} OxfordPets} 
& Prompt Tuning
& \textbf{87.89}
& \textbf{84.62}
& \textbf{81.20}
& \textbf{73.13} \\ \hline

& Full-Prompt-Tuning
& 72.36
& 63.14
& 55.29
& 38.69 \\
& CLS-Tuning
& 72.07
& 63.91
& 56.97
& 41.73\\
\rowcolor{tabhighlight}
\multirow{-3}{*}{\cellcolor{white} Food101} 
& Prompt Tuning 
& \textbf{76.99}
& \textbf{73.63}
& \textbf{71.07}
& \textbf{64.30} \\ \hline

& Full-Prompt-Tuning
& 62.80
& 55.50
& 49.01
& 34.66 \\
& CLS-Tuning
& 62.78
& 56.15
& 48.46
& 35.43\\
\rowcolor{tabhighlight}
\multirow{-3}{*}{\cellcolor{white} DTD} 
& Prompt Tuning 
& \textbf{62.86}
& \textbf{58.90}
& \textbf{53.62}
& \textbf{46.19}\\ \hline

& Full-Prompt-Tuning
& 73.02
& 64.31
& 57.11
& 40.42 \\
& CLS-Tuning
& 72.73
& 65.64
& 58.91
& 44.55\\
\rowcolor{tabhighlight}
\multirow{-3}{*}{\cellcolor{white} UCF101} 
& Prompt Tuning 
& \textbf{73.12}
& \textbf{68.73}
& \textbf{67.66}
& \textbf{60.93}\\

\bottomrule
\end{tabular}
\caption{Comparison of transfer performance at incremental noise rates between different prompt designs.}
\label{tbl:prompt_comp}
\end{table}

Table~\ref{tbl:classifier_comp} compares the various models on four datasets under different levels of label noise. 
The linear classifier with CLIP initialization (Classifier-C) outperformed random initialization across all levels of noise. This shows that CLIP class embeddings provide a strong initialization for few-shot learning. Furthermore, although both Classifiers degrade considerably with high noise ratios, the CLIP initialization is also more robust to noise. 
As for TEnc-FT, it achieved competitive performance at zero noise rates, but its accuracy also dropped significantly as the noise rate increased. This highlights (unsurprisingly) that the highly expressive CLIP text encoder can easily overfit to the noisy labels. 
Finally, Prompt Tuning outperformed all alternative strategies across all noise rates. The advantage of prompt tuning was especially large for high noise levels. These observations confirm that (a) the text encoder is essential for providing a strong but informative regularization of the text embeddings to combat noisy inputs (Prompt Tuning v.s. classifiers); and (b) the text encoder should be fixed to prevent overfitting (Prompt Tuning v.s. TEnc-FT).

\begin{figure*}[t]
 \begin{subfigure}{0.5\columnwidth}
     \includegraphics[width=\columnwidth]{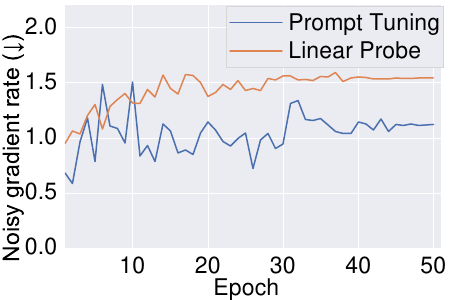}
     \caption{OxfordPets}
     \label{fig:gradient_oxfordpets}
 \end{subfigure}
 \hfill
 \begin{subfigure}{0.5\columnwidth}
     \includegraphics[width=\columnwidth]{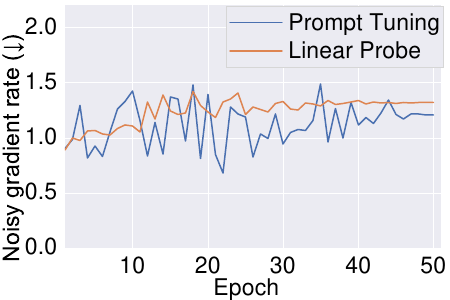}
     \caption{Food101}
     \label{fig:gradient_Food101}
 \end{subfigure}
  \hfill
  \begin{subfigure}{0.5\columnwidth}
     \includegraphics[width=\columnwidth]{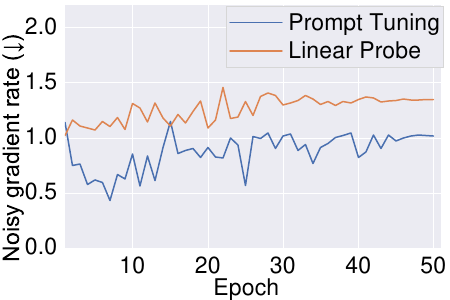}
     \caption{DTD}
     \label{fig:gradient_DTD}
 \end{subfigure}
  \hfill
  \begin{subfigure}{0.5\columnwidth}
     \includegraphics[width=\columnwidth]{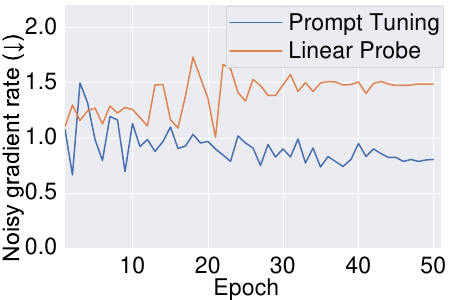}
     \caption{UCF101}
     \label{fig:gradient_UCF101}
 \end{subfigure}
 \caption{We assess the ability of both methods to suppress noisy gradients by evaluating their noisy-to-clean gradient norm ratio (noisy gradient rate). This ratio is determined by taking the L2 norm of gradients with respect to the learnable parameters, which we compute by feeding 64 clean samples and 64 noisy samples to the model during each training epoch.  Specifically, we train the models on data with a 50\% noise rate. Results on four datasets show that Prompt Tuning achieves a lower noisy gradient rate compared to Linear Probe, indicating its superior ability to suppress noisy gradients.}
 \label{fig:gradient}
\end{figure*}

\noindent \textbf{Effectiveness of prompt.} The previous experiment showed that the class embeddings generated by CLIP pre-trained text encoder plays a critical role in noise robustness. 
Next, we keep the text encoder fixed, and attempt to answer another question: \textit{Which components of the prompt provide noise robustness to prompt tuning?}

We hypothesize that the {{classname}} token $\mathbf{w}_c$ provides a strong regularization to the model, since it is leveraged by the text encoder to encode relationships between the different visual concepts (e.g.~how similar or different classes are from each other). Respecting this structure could help the model avoid fitting noisy data during training. To verify our hypothesis, we assess the noise robustness of two additional models:
\begin{description}
    \setlength\itemsep{0.05em}
    \item[Full Prompt Tuning] Learns the {{classname}} token jointly with the original learnable tokens (see Figure~\ref{fig:PT_ablation_TEnc_Prompt} (e)).
    \item[CLS Tuning] Adopts a \textit{fixed} template prompt ``A photo of a $<$CLS$>$'' and optimizes only the classname token (see Figure~\ref{fig:PT_ablation_TEnc_Prompt}(f)).
\end{description}

Table~\ref{tbl:prompt_comp} shows the analysis on four dataset for different noise levels. Compared to prompt-tuning, which optimizes only learnable tokens shared across all classes, both CLS-Tuning and Full-Prompt-Tuning models struggle at high noise rate. Even when the training data is clean, learning the {{classname}} tokens produces worse performance on two of the four datasets (OxfordPets and Food101).
This analysis validates our assumption that the fixed {{classname}} token is indeed a critical regularization for the prompt tuning. Learnable classname tokens can be fitted to the noisy training data, perturbing the class embeddings and leading to worse performance.

\subsection{Prompt Tuning Suppresses Noisy Gradients}
\label{sec:supress}

The previous section provided clear evidence of the robustness of the prompt tuning framework in comparison to other alternatives.
These findings suggest that, by learning only shared prompt tokens, prompt tuning focuses better on clean samples than noisy samples. In other words, prompt tuning can suppress gradient updates from noisy samples, while aggregating gradients from clean samples.
To verify this hypothesis, we measure the gradients with respect to the learnable parameters of both CLIP prompt tuning and linear probing using 50\% noise rate. Specifically, we measure the ratio between the gradient norm induced by noisy samples and that induced by clean samples. A ratio above one indicates that noisy samples play a bigger role in the optimization than clean samples.

Figure \ref{fig:gradient} shows the noisy-to-clean gradient norm ratio as models are trained on four datasets. As can be seen, prompt tuning displays significantly lower ratios than linear probing. This indicates that noisy samples play a comparatively small role with prompt tuning compared to linear probes.
This property likely arises from the highly constrained prompt tuning optimization, which restricting the model to fit the noisy labels.

\begin{figure}[t]
\centering
\begin{subfigure}{0.80\columnwidth}
    \includegraphics[width=\columnwidth]{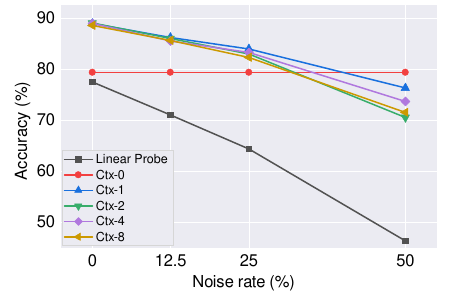}
    \caption{OxfordPets}
    \label{fig:ctx_length_OxforPets}
\end{subfigure}
\hfill
\begin{subfigure}{0.80\columnwidth}
    \includegraphics[width=\columnwidth]{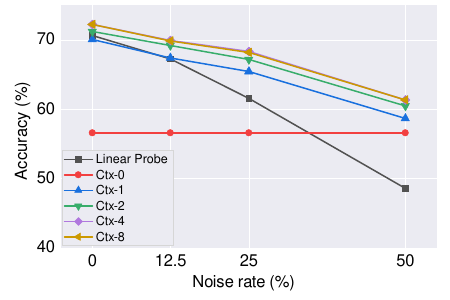}
    \caption{UCF101}
    \label{fig:ctx_length_Caltech101}
\end{subfigure}
\caption{Investigation on noise robustness of prompt tuning accompanied by various context lengths. Ctx-$x$ denotes the model with $x$ learnable tokens. }
 \label{fig:ctx_length}
\end{figure}

\setlength{\tabcolsep}{3pt}
\begin{table}[t]
\small
\centering
\begin{tabular}{ll|cccc}
\toprule
\multirow{2}{*}{\bf Dataset} &
\multirow{2}{*}{\begin{tabular}[c]{@{}c@{}} \bf Method \end{tabular}}
&\multicolumn{4}{c}{\bf Noise rate} \\
\multicolumn{2}{l|}{}                                 
& 0 & 12.5 
& 25
& 50 \\ \cline{3-6} \hline
\multirow{2}{*}{ImageNet} 
& RN50-PT
& 62.83
& 61.98
& 60.60
& 57.97\\
& ViT-B/32-PT
& 66.48
& 65.82
& 64.50
& 61.75 \\ \hline
\multirow{2}{*}{Caltech101} 
& RN50-PT
& 90.65
& 82.51
& 78.70
& 70.13\\ 
& ViT-B/32-PT
& 93.63
& 90.34
& 84.99
& 77.16 \\ \hline
\multirow{2}{*}{OxfordPets} 
& RN50-PT
& 87.89
& 84.62
& 81.20
& 73.13 \\
& ViT-B/32-PT
& 89.10
& 86.59
& 83.65
& 75.50 \\ \hline
\multirow{2}{*}{Flowers102} 
& RN50-PT
& 92.57
& 86.85
& 81.73
& 71.80 \\
& ViT-B/32-PT
& 93.26
& 87.90
& 85.34
& 72.83 \\ \hline
\multirow{2}{*}{Food101} 
& RN50-PT
& 76.99
& 73.63
& 71.07
& 64.30 \\
& ViT-B/32-PT
& 80.16
& 77.60
& 76.06
& 68.77 \\ \hline
\multirow{2}{*}{FGVCAircraft} 
& RN50-PT
& 27.13
& 25.07
& 23.34
& 19.05 \\
& ViT-B/32-PT
& 28.37
& 27.57
& 25.47
& 19.57 \\ \hline
\multirow{2}{*}{DTD} 
& RN50-PT
& 62.86
& 58.90
& 53.62
& 46.19 \\
& ViT-B/32-PT
& 64.88
& 59.57
& 57.09
& 45.22 \\ \hline
\multirow{2}{*}{UCF101} 
& RN50-PT
& 73.12
& 68.73
& 67.66
& 60.93 \\
& ViT-B/32-PT
& 78.12
& 75.97
& 72.83
& 65.75 \\ 

\bottomrule
\end{tabular}
\caption{Noise robustness of prompt tuning (PT) with ResNet50 or ViT-B/32 as the image encoder on eight datasets.}
\label{tbl:PT_on_more_datasets}
\end{table}



\subsection{Generalization Across Model Architectures}
Previous sections have focused on four datasets (OxfordPets, Food101, DTD, and UCF101) and a ResNet-50 image encoder. We now show that these findings generalize across model architectures and datasets.

\noindent\textbf{Context length.} We first assess the noise robustness of prompt tuning with increasing numbers of learnable tokens. We also evaluate a baseline without any learnable tokens by directly feeding the classname into the model (denoted as Ctx-0). Figure \ref{fig:ctx_length} shows that the optimal context lenght is dataset dependent, but all context lengths achieve superior performance compared to traditional linear probing. Ctx-0 outperforms some prompt tuning variants under large noise rates at 50\%, suggesting fixed prompts may be a good choice when the labeling noise is too strong on the downstream task.

\noindent\textbf{Image encoders.} To validate whether the noise robustness of prompt tuning is backbone-agnostic, we also assess CLIP with ViT-B/32 for prompt tuning (denoted ViT-B/32-PT). Table \ref{tbl:PT_on_more_datasets} shows the comparison with RN50-PT. ViT-B/32-PT outperforms RN50-PT under most settings. Moreover, both methods do not suffer from a large performance drop and maintain competitive accuracy at high noise rates.


\setlength{\tabcolsep}{5pt}
\begin{table}[t]
\centering
\small
\begin{tabular}{ll|cc}
\toprule
\bf Dataset & \bf Method & \bf Random & \bf Confusion \\ \cline{1-4}
\multirow{2}{*}{OxfordPets}
& Linear Probe  & 46.42\std{0.88} & 41.39\std{1.87} \\
& Prompt Tuning & 73.13\std{3.76} & 66.55\std{2.02} \\
\hline
\multirow{2}{*}{Food101} 
& Linear Probe & 42.63\std{0.89} & 37.71\std{0.52} \\
& Prompt Tuning & 64.30\std{2.58} & 63.93\std{1.45} \\
\hline
\multirow{2}{*}{DTD}
& Linear Probe & 42.29\std{2.12}  & 37.69\std{1.70} \\
& Prompt Tuning & 46.19\std{2.12}  & 45.76\std{1.23} \\
\hline
\multirow{2}{*}{UCF101}
& Linear Probe & 54.05\std{1.19}  & 50.90\std{1.45} \\
& Prompt Tuning & 60.93\std{0.94} & 59.11\std{0.70} \\
\bottomrule
\end{tabular}
\caption{The impact of random and confusion label noise at a 50\% noise rate on Linear Probing and Prompt Tuning strategies.}
\label{tbl:noise_type_ablation}
\end{table}

\subsection{Robustness to Correlated Label Noise}
So far, we assumed white label noise (i.e., noisy labels are uniformly drawn from the label space). However, label noise produced by either human annotators or machine-generated pseudo labels often displays correlations between similar visual concepts.
For example, UPL~\cite{Huang2022UPL} observed that pre-trained CLIP prefers some classes over others during zero-shot transfer. Inspired by this observation, we examine whether CLIP inherent preferences affect the performance of prompt tuning when confronted with CLIP-generated label noise. 

\begin{figure}[t]
    \includegraphics[width=1.0\columnwidth]{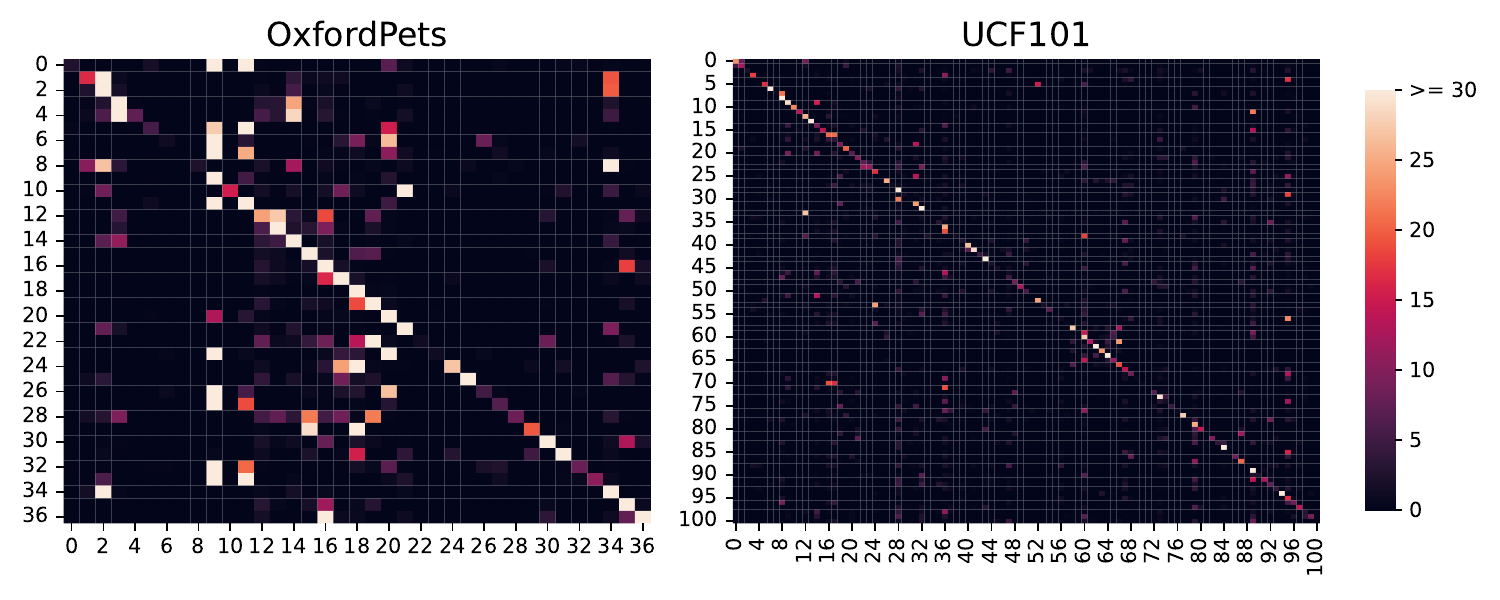}
    \caption{Confusion matrix generated by averaging the zero-shot performance over 100 runs using random prompt tokens.}
    \label{fig:confusion}
\end{figure}

We begin by measuring the confusion matrix of CLIP zero-shot predictions with \emph{randomly initialized} learnable tokens on the OxfordPets and UCF101 datasets (see Figure \ref{fig:confusion}). 
Next, we introduce a challenging type of label noise, named \emph{Confusion} noise, where each mislabeled sample is labeled as the incorrect class that is most favored by zero-shot CLIP. 
Finally, we examine the transfer performance of prompt tuning with both random and confusion noise at a 50\% noise rate. Table~\ref{tbl:noise_type_ablation} presents the results on four datasets. As can be seen, confusion noise presents a bigger challenge to transfer learning, leading to larger degradation of classification accuracy at high noise ratios compared to random noise.
Such degradation is visible both for prompt tuning and linear probes. However, among the two, prompt tuning still achieves the best overall performance, providing further evidence for its robustness even to more challenging types of noise.

\begin{figure*}[t]
    \centering
    \includegraphics[width=0.80\textwidth]{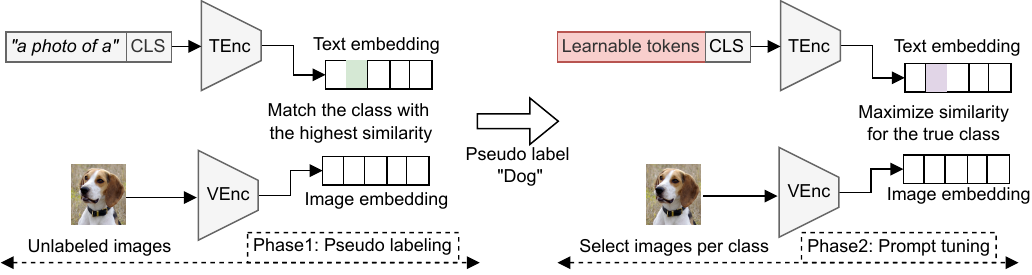}
    \caption{The pipeline of unsupervised prompt tuning. It consists of two main phases: Pseudo labeling and Prompt tuning. To begin, we generate pseudo labels for target datasets by utilizing CLIP with a template prompt for zero-shot transfer. Next, we randomly select samples per class from the pseudo labels for subsequent training. Finally, we optimize the learnable prompt representation using the selected pseudo-labeled samples.}
    \label{fig:upl_diagram}
\end{figure*}

\section{Application to Unsupervised Prompt Tuning}
\label{sec:UPT}
Prior work UPL~\cite{Huang2022UPL} demonstrated that unsupervised prompt tuning can outperform the transfer performance of zero-shot transfer based on CLIP. However, UPL does not fully utilize the noise robustness of prompt tuning. 

\noindent\textbf{Baseline UPL.} UPL~\cite{Huang2022UPL} proposed a framework to adapt CLIP for downstream tasks without any labeled images. An overview of the framework is shown in Figure \ref{fig:upl_diagram}. This framework is divided into two phases. 
In phase 1, UPL leverages pre-trained CLIP to generate pseudo labels for unlabeled images. 
Then, in phase 2, a set of $K$ pseudo-labels are chosen to optimize the learnable tokens through the typical prompt-tuning optimization process (described in CoOp~\cite{Zhou2022CoOp}). 
To increase the quality of training examples, UPL ranks all pseudo-labeled images based on their confidence score (Eq.~\ref{eq:cos_sim}) and selects the $K$ most confident samples per class.
Furthermore, inspired by prompt ensembling in CLIP~\cite{Radford2021CLIP}, UPL improved transfer performance by ensembling multiple predictions generated by models with different learnable prompts. 

\noindent\textbf{Robust UPL.} 
In Section~\ref{sec:exp-robustness}, we showed that prompt tuning can be robust to noisy labels. Furthermore, we showed that prompt tuning robustness can be further strengthened using the generalized cross-entropy loss (GCE). Given these observations, we propose to perform unsupervised prompt tuning by 
1) randomly sample training samples  and 2)  optimizing the prompt with the robust GCE loss.
Random sampling has two effects. On the one hand, it increases the diversity of training samples which benefits learning. On the other hand, it increases the amount of label noise. However, we expect the label noise to be tolerable by our robust prompt tuning framework.

\noindent\textbf{Experimental Settings.} We experiment with the unsupervised prompt tuning following the same training setting of Section \ref{sec:exp-robustness}. Pseudo-labels are generated by CLIP zero-transfer with ResNet50 image encoder. We follow the prompt engineering used by CLIP. There are three types of hand-crafted prompts, with more details listed in the supplementary material.
$K$ is set to 16 in all experiments. During the inference stage, we employ the ensemble-average approach following UPL~\cite{GenXEnt} to generate predictions combining the outputs of four distinct models. Each model has a distinct learnable prompt that was initialized with a unique random seed.

\noindent\textbf{Experimental Results.}
We compared UPL~\cite{Huang2022UPL} and the proposed Robust UPL on a diverse set of visual tasks, including generic object classification, fine-grained recognition, and texture identification. We also assessed Robust UPL using both a cross-entropy (CE) and generalized cross-entropy (GCE) losses. Table \ref{tbl:top16vsrandom} shows that all three unsupervised prompt tuning methods can improve transfer learning over zero-shot predictions, at no additional labeling cost. Among the three methods, Robust UPL trained under GCE loss obtains the best performance on average. We highlight once again that Robust UPL randomly samples pseudo labeled images for training, instead of using high-confidence samples as in UPL. As a result, UPL training pseudo-labels are less diverse, but have less noise. For example, the pseudo-labels used to train UPL on Caltech were 93\% correct, while the pseudo-labels used to train Robust UPL were only 83\% correct. Nevertheless, these errors did not harm final performance of Robust UPL; on the contrary,  learning from a more diverse set, while being robust to the noise enhanced prompt tuning.


\setlength{\tabcolsep}{5pt}
\begin{table}[t]
\small
\centering
\begin{tabular}{l|cccc}
\toprule
\multirow{2}{*}{\bf Dataset} & \multirow{2}{*}{\bf 0-Shot} & \multirow{2}{*}{\bf UPL~\cite{Huang2022UPL}} & \multicolumn{2}{c}{\bf Robust UPL}  \\ 
& & & CE & GCE\\ \hline
ImageNet    &58.18& 60.22 & 61.11 & \bf62.14 \\ 
Caltech101  &86.29& \bf 90.10 & 87.14 & 88.07\\
OxfordPets  &85.77& 87.60 & 86.89 & \bf 87.71\\
Flowers102  &66.14& 69.31 & 70.04 & \bf 70.52 \\
Food101     &77.31& 77.30 & 77.84 & \bf 78.51 \\
FGVCAircraft &\bf 17.28& 15.93 & 16.35 & 16.29 \\
DTD         &42.32& 37.47 & 44.80 & \bf 46.69 \\
UCF101      &61.46& 65.00 & 66.01 & \bf 67.12 \\ 
\bottomrule
\end{tabular}
\vspace{5pt}
\caption{Comparison between CLIP zero-shot classification and three strategies for unsupervised prompt tuning: UPL~\cite{Huang2022UPL}, and our robust UPL framework trained with cross-entropy and generalized cross-entropy losses.}
\label{tbl:top16vsrandom}
\end{table}

\section{Conclusion}
\label{sec:conclusion}
In this paper, we provide a comprehensive study on the robustness to label noise of prompt tuning large vision-language models (namely, CLIP). Through a series of experiments, we demonstrated that the noise robustness of prompt tuning can be attributed to the structure imposed on class embeddings by CLIP's pre-trained text encoder. We further demonstrate that prompt tuning can ease overfitting to mislabeled samples by reducing the gradients induced by label noise. 
We extensively experimented with different model configurations such as backbones and context length, obtaining consistent results and conclusions. 
Finally, inspired by our findings, we presented a new robust unsupervised prompt tuning approach that favors diversity over correct predictions, to improve the transfer performance.


{\small
\bibliographystyle{ieee_fullname}
\bibliography{main_arxiv}
}

\clearpage
\appendix 
\section{Extended Experimental Results}

\noindent\textbf{Robustness Attribution.}
The findings presented in Table~\ref{tbl:supp_classifier_comp} indicate that the reason behind the noise robustness observed in prompt tuning can be attributed to the structured form imposed on class embeddings by CLIP's pre-trained text encoder.

\setlength{\tabcolsep}{3pt}
\begin{table}[ht]
\small
\centering
\begin{tabular}{cl|cccccc}
\toprule
\multirow{2}{*}{\bf Dataset} &
\multirow{2}{*}{\begin{tabular}[c]{@{}c@{}} \bf Method \end{tabular}}
&\multicolumn{4}{c}{\bf Noise rate} \\ 
\multicolumn{2}{l|}{}                                 
& 0 & 12.5 & 25 & 50 \\ \cline{3-6} \hline

& Classifier-R
& 74.82
& 64.10
& 55.96
& 36.63 \\
& Classifier-C
& 81.47
& 70.29
& 61.87
& 44.21\\
& TEnc-FT
& 84.38
& 70.73
& 61.11
& 41.21 \\
\rowcolor{tabhighlight}
\multirow{-4}{*}{\cellcolor{white} OxfordPets} 
& Prompt Tuning
& \textbf{87.89}
& \textbf{84.62}
& \textbf{81.20}
& \textbf{73.13} \\ \hline

& Classifier-R
& 88.19
& 74.48
& 61.14
& 42.68 \\
& Classifier-C
& 89.94
& 77.04
& 63.81
& 45.96\\
& TEnc-FT
& \textbf{90.75}
& 76.67
& 62.76
& 46.45 \\
\rowcolor{tabhighlight}
\multirow{-4}{*}{\cellcolor{white} Caltech101} 
& Prompt Tuning
& 90.65
& \textbf{82.51}
& \textbf{78.70}
& \textbf{70.13} \\ \hline

& Classifier-R
& 83.40
& 69.58
& 60.85
& 37.74 \\
& Classifier-C
& 94.11
& 80.26
& 69.18
& 47.55\\
& TEnc-FT
& \textbf{94.62}
& 80.91
& 70.83
& 49.54 \\
\rowcolor{tabhighlight}
\multirow{-4}{*}{\cellcolor{white} Flowers102} 
& Prompt Tuning
& 91.71
& \textbf{86.24}
& \textbf{81.92}
& \textbf{71.80} \\ \hline

& Classifier-R
& 63.80
& 54.66
& 46.23
& 28.97 \\
& Classifier-C
& 69.36
& 60.46
& 51.85
& 34.37\\
& TEnc-FT
& 71.30
& 61.60
& 52.64
& 34.74 \\
\rowcolor{tabhighlight}
\multirow{-4}{*}{\cellcolor{white} Food101} 
& Prompt Tuning
& \textbf{76.99}
& \textbf{73.63}
& \textbf{71.07}
& \textbf{64.30} \\ \hline

& Classifier-R
& 30.38
& 24.84
& 20.89
& 13.32 \\
& Classifier-C
& 34.46
& \textbf{29.97}
& \textbf{25.91}
& 17.36\\
& TEnc-FT
& \textbf{35.30}
& 29.66
& 25.31
& 17.42 \\
\rowcolor{tabhighlight}
\multirow{-4}{*}{\cellcolor{white} FGVCAircr} 
& Prompt Tuning
& 27.13
& 25.07
& 23.34
& \textbf{19.05} \\ \hline

& Classifier-R
& 48.02
& 44.30
& 40.32
& 30.10 \\
& Classifier-C
& \textbf{63.83}
& 57.14
& 50.36
& 34.86\\
& TEnc-FT
& 63.61
& 55.47
& 48.21
& 33.12 \\
\rowcolor{tabhighlight}
\multirow{-4}{*}{\cellcolor{white} DTD} 
& Prompt Tuning
& 62.86
& \textbf{58.90}
& \textbf{53.62}
& \textbf{46.19} \\ \hline

& Classifier-R
& 67.16
& 58.33
& 50.34
& 31.07 \\
& Classifier-C
& 71.87
& 64.12
& 54.79
& 38.01\\
& TEnc-FT
& \textbf{73.74}
& 64.52
& 56.10
& 37.88 \\
\rowcolor{tabhighlight}
\multirow{-4}{*}{\cellcolor{white} UCF101} 
& Prompt Tuning
& 73.12
& \textbf{68.73}
& \textbf{67.66}
& \textbf{60.93} \\ 

\bottomrule
\end{tabular}
\caption{Comparison of transfer performance at incremental noise rates between different variants.}
\label{tbl:supp_classifier_comp}
\end{table}

\setlength{\tabcolsep}{2pt}
\begin{table}[h]
\small
\centering
\begin{tabular}{ll|cccccc}
\toprule
\multirow{2}{*}{Dataset} &
\multirow{2}{*}{\begin{tabular}[c]{@{}c@{}} Method \end{tabular}}
&\multicolumn{4}{c}{Noise rate} \\ \cline{3-6} 
\multicolumn{2}{l|}{}                                 
& 0 & 12.5 
& 25
& 50 \\ \cline{3-6} \hline
& Full-Prompt-Tuning
& 85.39
& 74.00
& 68.66
& 50.50\\
& CLS-Tuning
& 85.04
& 77.02
& 71.03
& 53.15 \\
\rowcolor{tabhighlight}
\multirow{-3}{*}{\cellcolor{white} OxfordPets} 
& Prompt Tuning
& \textbf{87.89}
& \textbf{84.62}
& \textbf{81.20}
& \textbf{73.13} \\ \hline

& Full-Prompt-Tuning
& 89.21
& 74.20
& 61.26
& 45.92 \\
& CLS-Tuning
& 89.13
& 76.84
& 62.27
& 48.64\\
\rowcolor{tabhighlight}
\multirow{-3}{*}{\cellcolor{white} Caltech101} 
& Prompt Tuning
& \textbf{90.65}
& \textbf{82.51}
& \textbf{78.70}
& \textbf{70.13}\\ \hline

& Full-Prompt-Tuning
& \textbf{93.93}
& 83.58
& 77.00
& 59.52 \\
& CLS-Tuning
& 93.47
& 84.19
& 78.74
& 61.79\\
\rowcolor{tabhighlight}
\multirow{-3}{*}{\cellcolor{white} Flowers102} 
& Prompt Tuning
& 91.71
& \textbf{86.24}
& \textbf{81.92}
& \textbf{71.80}\\ \hline

& Full-Prompt-Tuning
& 72.36
& 63.14
& 55.29
& 38.69 \\
& CLS-Tuning
& 72.07
& 63.91
& 56.97
& 41.73\\
\rowcolor{tabhighlight}
\multirow{-3}{*}{\cellcolor{white} Food101} 
& Prompt Tuning
& \textbf{76.99}
& \textbf{73.63}
& \textbf{71.07}
& \textbf{64.30} \\ \hline

& Full-Prompt-Tuning
& \textbf{32.28}
& \textbf{28.16}
& \textbf{24.67}
& 16.76 \\
& CLS-Tuning
& 30.84
& 27.86
& 24.51
& 17.63\\
\rowcolor{tabhighlight}
\multirow{-3}{*}{\cellcolor{white} FGVCAircraft} 
& Prompt Tuning
& 27.13
& 25.07
& 23.34
& \textbf{19.05}\\ \hline

& Full-Prompt-Tuning
& 62.80
& 55.50
& 49.01
& 34.66 \\
& CLS-Tuning
& 62.78
& 56.15
& 48.46
& 35.43\\
\rowcolor{tabhighlight}
\multirow{-3}{*}{\cellcolor{white} DTD} 
& Prompt Tuning
& \textbf{62.86}
& \textbf{58.90}
& \textbf{53.62}
& \textbf{46.19}\\ \hline

& Full-Prompt-Tuning
& 73.02
& 64.31
& 57.11
& 40.42 \\
& CLS-Tuning
& 72.73
& 65.64
& 58.91
& 44.55\\
\rowcolor{tabhighlight}
\multirow{-3}{*}{\cellcolor{white} UCF101} 
& Prompt Tuning
& \textbf{73.12}
& \textbf{68.73}
& \textbf{67.66}
& \textbf{60.93}\\

\bottomrule
\end{tabular}
\caption{Comparison of transfer performance at incremental noise rates between different prompt designs.}
\label{tbl:supp_prompt_comp}
\end{table}

Table~\ref{tbl:supp_prompt_comp} validates two observations: (a) the significance of the text encoder in offering robust regularization of the text embeddings to tackle noisy inputs (Prompt Tuning versus classifiers); and (b) the necessity of fixing the text encoder to prevent overfitting (Prompt Tuning versus TEnc-FT).

 \noindent\textbf{Robustness to Correlated Label Noise.}
 Table~\ref{tbl:supp_noise_type_ablation} shows that transfer learning faces a greater challenge with confusion noise, resulting in a great decline in classification accuracy at higher noise ratios as opposed to random noise. This decline is evident in both prompt tuning and linear probes. The robustness of prompt tuning is evident in its ability to outperform linear probes, even when faced with more challenging noise types.

\begin{figure*}[t]
  \begin{center}
    \includegraphics[width=1.0\textwidth]{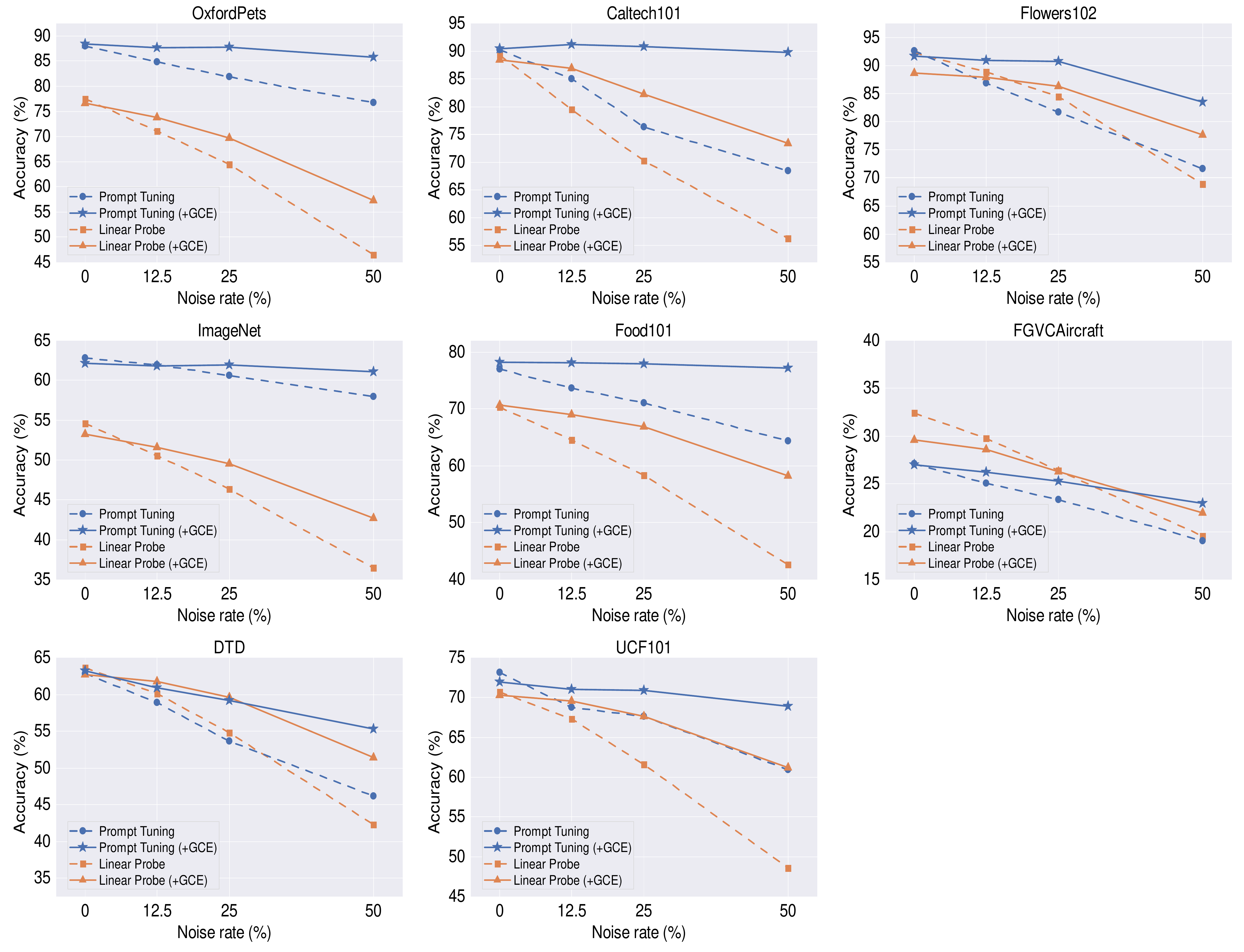}
  \end{center}
  \caption{Incorporating the generalized cross-entropy (GCE)~\cite{GenXEnt} loss with Prompt Tuning and Linear Probe methods, originally trained using cross-entropy, can enhance their noise robustness. At high noise rates. PromptTuning(+GCE) consistently and significantly outperforms other approaches on all datasets.}
  \label{fig:supp_GCE_comp}
\end{figure*}

\begin{figure}[t]
\centering
 \begin{subfigure}{0.8\columnwidth}
     \includegraphics[width=1.0\columnwidth]{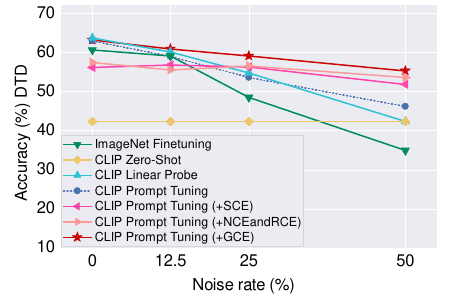}
     \label{fig:rubust_losses_DTD}
 \end{subfigure}
 \hfill
 \begin{subfigure}{0.8\columnwidth}
     \includegraphics[width=1.0\columnwidth]{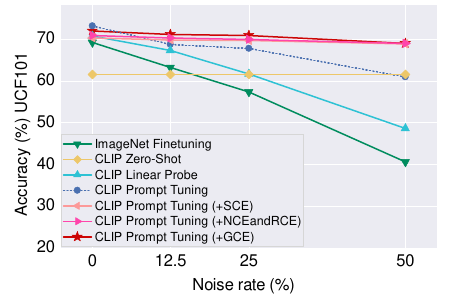}
     \label{fig:rubust_losses_ucf101}
 \end{subfigure}
 \vspace*{-8mm}
 \caption{Combination of traditional transfer learning and prompt tuning approaches, with three robust loss functions.}
 \label{fig:rubust_losses}
\end{figure}

\noindent\textbf{Integration with Noise-Robust Losses.}
We examine the effectiveness of a robust loss function
applied to prompt tuning with noisy labels. In this study, We adopt  Generalized Cross Entropy (GCE)~\cite{GenXEnt} as a representative of robust loss functions for noise-robust learning. Specifically, cross-entropy loss in Eq.~2 is replaced with GCE loss during the training process. Figure~\ref{fig:supp_GCE_comp} shows results of applying GCE loss to prompt tuning and linear probing for CLIP’s vision encoder. We observe that both transfer learning methods obtain an improvement of noise robustness by training with GCE loss. In particular, prompt tuning further enhances its inherent noise robustness. This outcome suggests that prompt tuning offers great applicability to couple with existing noise-robust loss functions. In addition to GCE, Figure~\ref {fig:rubust_losses} shows two additional robust loss functions: Symmetric Cross Entropy (SCE)~\cite{wang2019symmetric} and Normalized Cross Entropy (NCE) combined with a Reverse Cross Entropy (RCE)~\cite{ma2020normalized_loss}. Both losses also improve the noise robustness of prompt tuning, but GCE still achieves slightly better performance.

\setlength{\tabcolsep}{9pt}
\begin{table}[h]
\small
\centering
\begin{tabular}{l|l|c}
\toprule
Dataset &  Method  & 0-Shot\\ \hline
OxfordPets &  Random Prompt  & 40.93\std{11.19}
 \\ \hline
Caltech101 &  Random Prompt  & 59.65\std{9.56}
 \\ \hline
 Flowers102 &  Random Prompt   & 14.86\std{8.40}
 \\ \hline
 Food101 &  Random Prompt  & 34.63\std{11.10}
 \\ \hline
 FGVCAircraft &  Random Prompt   & 3.43\std{1.97}
 \\ \hline
 DTD &  Random Prompt  & 21.20\std{3.51}
 \\ \hline
 UCF101 &  Random Prompt   & 32.93\std{5.48}
 \\ 
\bottomrule
\end{tabular}
\caption{CLIP zero-shot with random prompts.}
\label{tbl:supp_zero-shot_random}
\end{table}

\setlength{\tabcolsep}{7pt}
\begin{table}[h]
\centering
\small
\begin{tabular}{ll|cc}
\toprule
\bf Dataset & \bf Method & \bf Random & \bf Confusion \\ \cline{1-4}
\multirow{2}{*}{OxfordPets}
& Linear Probe  & 46.42\std{0.88} & 41.39\std{1.87} \\
& Prompt Tuning & 73.13\std{3.76} & 66.55\std{2.02} \\
\hline
\multirow{2}{*}{Caltech101}
& Linear Probe  & 56.24\std{1.96} & 56.25\std{6.92} \\
& Prompt Tuning & 70.13\std{3.76} & 70.86\std{1.83} \\
\hline
\multirow{2}{*}{Flowers102}
& Linear Probe  & 68.92\std{0.76} & 45.94\std{0.69} \\
& Prompt Tuning & 71.80\std{1.00} & 69.63\std{1.31} \\
\hline
\multirow{2}{*}{Food101} 
& Linear Probe & 42.63\std{0.89} & 37.71\std{0.52} \\
& Prompt Tuning & 64.30\std{2.58} & 63.93\std{1.45} \\
\hline
\multirow{2}{*}{FGVCAircraft} 
& Linear Probe & 21.98\std{0.48} & 15.38\std{0.71} \\
& Prompt Tuning & 19.05\std{1.06} & 18.04\std{1.32} \\
\hline
\multirow{2}{*}{DTD}
& Linear Probe & 42.29\std{2.12}  & 37.69\std{1.70} \\
& Prompt Tuning & 46.19\std{2.12}  & 45.76\std{1.23} \\
\hline
\multirow{2}{*}{UCF101}
& Linear Probe & 54.05\std{1.19}  & 50.90\std{1.45} \\
& Prompt Tuning & 60.93\std{0.94} & 59.11\std{0.70} \\
\bottomrule
\end{tabular}
\caption{The impact of random and confusion label noise at a 50\% noise rate on Linear Probing and Prompt Tuning strategies.}
\label{tbl:supp_noise_type_ablation}
\end{table}

\section{Capacity of classifiers with random prompt tokens} 
Prompt tuning has a limited parameter space given by the length of the prompt tokens.
This parameter space is fundamentally different from that of a whole neural network and may present special properties related to the robustness of the model. 
Instead of updating the learnable prompts with noisy data, we evaluate the classifiers with \emph{random prompts }. Table~\ref{tbl:supp_zero-shot_random} summarizes the average zero-shot performance over 100 runs. Surprisingly, the results show that CLIP can achieve non-trivial zero-shot performance, even with random prompts. 
This indicates that as long as the classname token is provided to the pre-trained text encoder, CLIP is capable of computing non-trivial class embeddings for generic image classification.

\section{Unsupervised Prompt Tuning Settings} Pseudo-labels are generated by CLIP zero-transfer with ResNet50 image encoder. We follow the prompt engineering used by CLIP. There are three types of hand-crafted prompts: "A photo of a $<$label name$>$" for generic object datasets; "A photo of a $<$label name$>$, a type of $<$collective name$>$" for fine-grained object datasets (e.g., prompts for OxfordPets are appended "a type of dog" or "a type of cat"); and "$<$label name$>$ texture" for the DTD dataset.
$K$ is set to 16 in all experiments.

\end{document}